\documentclass[preprint,12pt]{elsarticle}

\usepackage{amssymb}
\usepackage{amsmath}

\usepackage{todonotes}

\usepackage{booktabs}
\usepackage{tabularx}
\usepackage{threeparttable}
\usepackage{array}
\usepackage{arydshln}
\usepackage{caption}
\usepackage{hyperref}

\journal{Nuclear Physics B}

\begin{document}

\begin{frontmatter}



\title{Colorimeter-Supervised Skin Tone Estimation from Dermatoscopic Images for Fairness Auditing} 


\author{Marin Ben\v{c}evi\'{c}\corref{cor1}}
\ead{marin.bencevic@ferit.hr}

\author{Kre\v{s}imir Romi\'{c}}
\author{Ivana Hartmann Toli\'{c}}
\author{Irena Gali\'{c}}

\cortext[cor1]{Corresponding author.}

\affiliation{organization={Faculty of Electrical Engineering, Computer Science and Information Technology Osijek, Josip Juraj Strossmayer University of Osijek},
            country={Croatia}}

\begin{abstract}
\textit{Background and Objective:} Neural network-based diagnosis of dermatoscopic images is increasingly used for clinical decision support, yet multiple studies report performance disparities across skin tones. Fairness auditing of these models is limited by the lack of reliable skin tone annotations in public dermatoscopy datasets. Prior work has estimated skin tone from images by selecting representative skin pixels. However, such approaches are unreliable because the relationship between physical skin color and image pixel values is complex. We therefore train neural networks to directly learn the mapping between dermatoscopic images and physical skin color using colorimeter measurements.

\textit{Methods:} We train neural networks to predict Fitzpatrick skin type via ordinal regression and the Individual Typology Angle via color regression. Models are trained using dermatologist in-person Fitzpatrick labels and colorimeter measurements as targets, with extensive pretraining on synthetic and real dermatoscopic and clinical images. Evaluation is performed on a cohort of 64 subjects balanced by skin tone. Fitzpatrick type agreement is assessed using linear-weighted Cohen's kappa, and Individual Typology Angle agreement is assessed using intraclass correlation (ICC3). We then perform inference on ISIC 2020 and MILK10k to characterize skin-tone distributions in representative public benchmarks.

\textit{Results:} Fitzpatrick type predictions show agreement slightly lower than that of human annotaters, achieving an overall linear-weighted Cohen's kappa of 52.98\% (95\% CI: 51.54, 54.40), compared to 66.08\% (95\% CI: 64.21, 67.91) for crowdsourced annotations. More reliably, Individual Typology Angle predictions show high concordance with colorimeter-derived values, with ICC3 of 93.88\% (95\% CI: 93.31, 94.40), compared to 98.38\% (95\% CI: 98.24, 98.50) for colorimeter measurements, outperforming previously published pixel-averaging methods. Applying these estimators to ISIC 2020 and MILK10k reveals a strong underrepresentation of darker skin tones, with fewer than 1\% of subjects estimated as Fitzpatrick types V and VI.

\textit{Conclusions:} Neural networks can provide fast, scalable skin-tone estimation from dermatoscopic images and substantially outperform approaches based on selecting representative pixel values. These estimators enable large-scale skin tone annotation of public dermatoscopy datasets, supporting more rigorous bias auditing and improved fairness evaluation of dermatoscopic classifiers.
\end{abstract}




\end{frontmatter}



\section{Introduction}

Deep learning systems for dermatoscopic image analysis are increasingly used as clinical decision support, particularly for triage and risk stratification of pigmented lesions. Alongside rapid progress in model accuracy, there is a growing body of evidence that dermatology AI systems can exhibit clinically relevant performance gaps across skin tones \cite{daneshjouDisparitiesDermatologyAI2022, grohEvaluatingDeepNeural2021, daneshjouLackTransparencyPotential2021}. In dermatoscopy specifically, auditing such bias is hindered by a practical limitation: commonly used public benchmarks typically lack reliable skin-tone annotations \cite{daneshjouLackTransparencyPotential2021}. As a result, fairness audits are often restricted to indirect proxies, such as estimating skin tone directly from image pixels using computer-vision heuristics \cite{bevanDetectingMelanomaFairly2022, kinyanjuiEstimatingSkinTone2019, liEstimatingImprovingFairness2021, bencevicUnderstandingSkinColor2024b}. However, recent evidence suggests that pixel values in dermatoscopic images have a complex and nonlinear relationship to physical skin tone, showing that previously published skin tone estimation methods could be unreliable \cite{weirEvaluatingSkinTone2025}. In contrast, our goal is to learn the complex mapping from dermatoscopic pixels to skin tone using a supervised neural network with physical colorimeter measurements and in-person Fitzpatrick type annotations as ground truth.

We choose to operationalize skin tone using the Fitzpatrick (FP) type scale and Individual Typology Angle (ITA) as they are most commonly used in dermatological practice. Fitzpatrick type is widely used and clinically familiar, but image-based labeling can be subjective \cite{daneshjouDisparitiesDermatologyAI2022, grohEvaluatingDeepNeural2021}. ITA, on the other hand, provides a continuous measure of pigmentation derived from CIE L\textsuperscript{*}a\textsuperscript{*}b\textsuperscript{*} color space:
\begin{equation}
\mathrm{ITA}(L^{*}, b^{*}) = \arctan\!\left(\frac{L^{*}-50}{b^{*}}\right)\frac{180}{\pi}.
\label{eq:ita}
\end{equation}
ITA has been widely used in image-based fairness audits \cite{kinyanjuiEstimatingSkinTone2019, liEstimatingImprovingFairness2021, bevanDetectingMelanomaFairly2022}, but its validity in dermatoscopy depends on how well pixel-derived estimates can track physical measurements under variable acquisition conditions \cite{weirEvaluatingSkinTone2025}.

In this paper, we address skin tone annotation by introducing a neural-network-based skin tone estimation pipeline for dermatoscopic (lesional and non-lesional) images that outputs both Fitzpatrick type and ITA. We formulate the Fitzpatrick prediction as ordinal regression and the ITA prediction as color regression. We train and evaluate using the MSKCC Skin Tone Labeling dataset \cite{weirEvaluatingSkinTone2025}, which provides in-person expert Fitzpatrick labels, image-based crowdsourced Fitzpatrick labels, as well as paired triplicate colorimeter measurements. We further leverage extensive pretraining on synthetic and real dermatoscopic images. We compare our results to crowdsourced human labels as well as the repeatability of multiple colorimeter measurements. Our Fitzpatrick estimator achieves agreement slightly lower than human raters, and our ITA predictions show high concordance with colorimeter-derived ITA, substantially exceeding previously used computer-vision baselines \cite{bevanDetectingMelanomaFairly2022, bencevicUnderstandingSkinColor2024b}.

Using these estimators, we annotate widely used public benchmarks and quantify the resulting skin-tone distributions. In particular, evaluate ISIC 2020 \cite{codellaSkinLesionAnalysis2019b} and MILK10k \cite{tschandlMILK10kHierarchicalMultimodal2026}, demonstrating strong underrepresentation of darker skin tones and supporting prior concerns about dataset composition \cite{daneshjouLackTransparencyPotential2021}. By releasing code and pretrained models as an open-source tool, we enable fast skin-tone annotation and reproducible bias audits on dermatoscopic datasets.

In short, our contributions are threefold:
\begin{enumerate}
    \item We present and publicly release a method for accurate ITA estimation from dermatoscopic images that significantly outperforms previously used computer-vision approaches, and we validate it against colorimeter measurements. To our knowledge, this is the first study to publish a neural network for ITA estimation of dermatoscopic images using colorimeter ground truth ITA values.
    \item We introduce an ordinal regression model for Fitzpatrick type prediction and evaluate agreement relative to human evaluators on dermatoscopic images.
    \item We provide an evaluation of skin tone distributions in commonly used dermatoscopic benchmarks and report severe underrepresentation of darker skin tones, including on ISIC 2020 \cite{codellaSkinLesionAnalysis2019b} and MILK10k \cite{tschandlMILK10kHierarchicalMultimodal2026}.
\end{enumerate}

The model and analysis code, necessary data, inference results, and trained models are available at \href{https://github.com/marinbenc/nn_colorimetry_dermatoscopy}{github.com/marinbenc/nn\_colorimetry\_dermatoscopy}.

\subsection{Related work}

One of the main challenges in quantifying this bias is the lack of ground truth skin tone information in commonly used datasets \cite{daneshjouLackTransparencyPotential2021}. To address this limitation, prior work has estimated skin tone directly from dermatoscopic image pixels using computer vision. For example, \citet{kinyanjuiEstimatingSkinTone2019} assessed the skin tone distribution of the ISIC 2019 dataset \cite{codellaSkinLesionAnalysis2019b} by computing the individual typology angle from the average color across non-lesional skin pixels. Other similar ITA estimation approaches have incorporated additional preprocessing or aggregation strategies, including white balancing \cite{liEstimatingImprovingFairness2021}, patch-based averaging \cite{bevanDetectingMelanomaFairly2022, corbinExploringStrategiesGenerate2023}, and k-means color clustering \cite{loaizaSKINTONEPROBLEM2020, bencevicUnderstandingSkinColor2024b}. However, these methods suffer from the fact that the values of pixels of skin on dermatoscopic images have a complex and nonlinear relationship to physical skin tone \cite{weirEvaluatingSkinTone2025}. 

More complex methods of modeling skin tone from pixels have been proposed, such as the one by \citet{khalkhaliMSTAISkinColor2025}, where skin tone is modeled as a probability density function instead of a single color value. This approach is able to better capture within-image variability of skin tone. Nevertheless, their approach is trained without a colorimeter ground truth and therefore remains sensitive to acquisition conditions, including lighting variation. In contrast, we estimate ITA using a neural network trained to learn the mapping between pixel values and physical color measurements.

Beyond continuous measures such as ITA, skin tone can also be represented categorically using established scales such as Fitzpatrick type \cite{fitzpatrickValidityPracticalitySunReactive1988a} or the Monk Skin Tone scale \cite{schumannConsensusSubjectivitySkin2024}. As such, skin tone estimation can be treated as an image classification problem. A small number of such datasets exist for clinical dermatology images \cite{grohEvaluatingDeepNeural2021, daneshjouDisparitiesDermatologyAI2022} and, more recently, for dermatoscopic images \cite{weirEvaluatingSkinTone2025, tschandlMILK10kHierarchicalMultimodal2026}. However, to our knowledge, there is a lack of studies evaluating the accuracy and reliability of neural network–based skin-tone classifiers in the dermatoscopic setting. We address this gap by training an ordinal regression model to classify skin tones using the Fitzpatrick scale.

\section{Methods}

We estimate skin tone from dermatoscopic images using two neural networks. The first is an ordinal regression model that predicts the three CIELAB color channels ($L^*$, $a^*$, $b^*$). The second is a classifier that predicts Fitzpatrick skin type (six classes). In this section, we will describe in detail the datasets used in our experiments and the construction of our neural network models.

\subsection{Data Description}

\begin{table*}[h]
\centering
\footnotesize
\begin{threeparttable}
\caption{An overview of used dermatological datasets. FP = Fitzpatrick type, ITA = Individual Typology Angle, MST = Monk Skin Tone.}
\label{tab:skin_tone_datasets}

\setlength{\tabcolsep}{5pt}
\renewcommand{\arraystretch}{1.15}

\begin{tabularx}{\textwidth}{@{}l l l X@{}}
\toprule
Dataset & Skin Tone Label & Size & Use in this work \\
\midrule

\multicolumn{4}{@{}l}{\textbf{Dermatoscopic}}\\
\addlinespace[2pt]
\hspace{0.6em}MSKCC \cite{weirEvaluatingSkinTone2025} &
FP; ITA\tnote{a} &
64 subjects; 4{,}878 images &
Training; Testing \\
\addlinespace
\hspace{0.6em}MILK-10k \cite{tschandlMILK10kHierarchicalMultimodal2026} &
$[0\mathrel{{.}\,{.}}\nobreak5]$\tnote{b} &
5{,}240 lesions &
Inference \\
\addlinespace
\hspace{0.6em}MRA-MIDAS \cite{chiouMRAMIDASMultimodalImage2024} &
FP &
796 subjects; 3{,}830 images &
Pre-training \\
\addlinespace
\hspace{0.6em}ISIC 2020 \cite{rotembergPatientcentricDatasetImages2021a} &
None &
33{,}126 images &
Inference \\
\addlinespace[4pt]

\addlinespace[2pt]
\multicolumn{4}{@{}l}{\textbf{Clinical}}\\
\addlinespace[2pt]
\hspace{0.6em}PAD-UFES-20 \cite{pachecoPADUFES20SkinLesion2020a} &
FP &
1{,}373 subjects; 2{,}298 images &
Pre-training \\
\addlinespace
\hspace{0.6em}SCIN \cite{wardCreatingEmpiricalDermatology2024} &
FP; MST\tnote{c} &
$>$10{,}000 images &
Pre-training \\
\addlinespace
\hspace{0.6em}Fitzpatrick17k \cite{grohEvaluatingDeepNeural2021} &
FP\tnote{c} &
16{,}577 images &
Pre-training \\
\addlinespace
\hspace{0.6em}DDI \cite{daneshjouDisparitiesDermatologyAI2022} &
FP &
570 subjects; 656 images &
Pre-training \\
\addlinespace[4pt]

\addlinespace[2pt]
\multicolumn{4}{@{}l}{\textbf{Synthetic}}\\
\addlinespace[2pt]
\hspace{0.6em}S-SYNTH \cite{kimSSYNTHKnowledgeBasedSynthetic2024} &
$[0, 1]$\tnote{d} &
10{,}000 images &
Pre-training \\

\bottomrule
\end{tabularx}

\begin{tablenotes}[flushleft]
\footnotesize
\item[a] Based on average triplicate CIELAB colorimeter measurements.
\item[b] Custom skin tone scale.
\item[c] Skin tone assessed from photographs, not from in-person visits.
\item[d] Melanin content is a continuous value input into the simulation„ pipeline.
\end{tablenotes}
\end{threeparttable}
\end{table*}

To support the experiments in this paper, we use a curated set of public dermatology datasets for three purposes: (i) pretraining, (ii) fine-tuning and evaluation, and (iii) inference to characterize skin tone distributions (Table~\ref{tab:skin_tone_datasets}). Our primary objective is skin tone estimation from \textit{dermatoscopic} images; nevertheless, because explicit skin-tone labels are more common in clinical photography, we also include clinical datasets (i.e., images of skin captured with cameras) in our broader data collection.

The MSKCC Skin Tone Labeling dataset \cite{weirEvaluatingSkinTone2025} is our primary dataset for fine-tuning and evaluation. It contains 4,800+ dermatoscopic and clinical close-up images of human skin (lesional and non-lesional sites) from 64 subjects, with multiple images per subject captured under varying dermatoscopic acquisition conditions. Crucially, it provides gold-standard skin tone information in two complementary forms: in-person expert Fitzpatrick labels for each subject and triplicate colorimeter measurements collected around the lesion (triplicate measurements at multiple nearby locations), enabling direct validation of both ordinal (Fitzpatrick) and continuous (ITA) skin tone estimators. We use MSKCC in a patient-level 5-fold cross-validation setup to avoid subject leakage between training and test splits.

In addition to the datasets used for training and evaluation, we run inference on widely used dermatoscopic benchmarks that lack Fitzpatrick or ITA labels to characterize their skin-tone distributions. This includes ISIC 2020 \cite{rotembergPatientcentricDatasetImages2021a}, a standard benchmark for melanoma detection that does not provide skin-tone annotations. We also evaluate MILK10k \cite{tschandlMILK10kHierarchicalMultimodal2026}, which includes both clinical and dermatoscopic images and provides a custom 0--5 skin-tone label distinct from Fitzpatrick type and ITA. These datasets are used for inference-only analysis and are not used during training.

\subsection{Data Preprocessing and Pretraining Data}

For pretraining, we use several clinical and dermatoscopic datasets with Fitzpatrick labels and train a neural network classifier (described below) directly on their provided annotations. These include PAD-UFES-20 \cite{pachecoPADUFES20SkinLesion2020a}, SCIN \cite{wardCreatingEmpiricalDermatology2024}, Fitzpatrick17k \cite{grohEvaluatingDeepNeural2021}, and MRA-MIDAS \cite{chiouMRAMIDASMultimodalImage2024}.

For the Diverse Dermatology Images dataset (DDI) \cite{daneshjouDisparitiesDermatologyAI2022}, which reports three grouped classes (FP I--II, III--IV, V--VI), we randomly assign each image to one of the corresponding subtypes (e.g., FP I--II mapped to FP I or FP II).

Finally, we additionally pretrain on a synthetic dataset generated with S-SYNTH \cite{kimSSYNTHKnowledgeBasedSynthetic2024}, which renders dermatoscopic images from controllable parameters including melanin content. We convert melanin values to Fitzpatrick labels by binning with empirically determined thresholds, following \cite{bencevicSkinColorMeasurement2025}.

During training and pretraining, each image is resized to $224\times224$ and normalized using ImageNet mean and standard deviation. 

\subsection{Neural Network Backbone and Pretraining}

Both models share the same EfficientNet-B0 backbone \cite{tanEfficientNetRethinkingModel2020a}, initialized with ImageNet weights. We pretrain the backbone using a Fitzpatrick ordinal regression head (described further in the text) on the union of all pretraining datasets listed in Table~\ref{tab:skin_tone_datasets}.

Backbone pretraining uses a batch size of 16 and a learning rate of $10^{-3}$ for up to 50 epochs, with early stopping (patience = 5 epochs) based on validation loss.

For Fitzpatrick classification and CIELAB-based ITA estimation, we perform 5-fold patient-level cross-validation, yielding five models per task. For inference on external datasets, we use an ensemble of the five models and report the mean prediction.

\subsection{Fitzpatrick type classification}

Once pretrained, the network is fine-tuned on the MSKCC Skin Tone Labeling \cite{weirEvaluatingSkinTone2025} dataset using a CORAL ordinal regression \cite{caoRankConsistentOrdinal2020} head. We found that ordinal regression, while being theoretically correct for the problem of skin tone estimation, also produces more reliable results, as shown in \ref{app1}. 

Pretraining is performed for up to 30 epochs using a learning rate of $10^{-4}$ and a batch size of 32, with early stopping (patience = 5 epochs) based on validation loss. We then fine-tune the network on the full MSKCC dataset, including clinical images; however, testing is restricted to dermatoscopic images only.

\subsection{ITA estimation}

Separately, for ITA estimation, the pre-trained backbone is fine-tuned with a regression head that outputs the three CIELAB channels ($L^{*}$, $a^{*}$, $b^{*}$). The ground truth values come from the MSKCC Skin Tone Labeling dataset \cite{weirEvaluatingSkinTone2025} \footnote{``Analysis of colorimeter vs.~image-extracted ITA in non-lesional sites.'' \href{https://doi.org/10.34970/962049}{https://doi.org/10.34970/962049}} and are computed as the average of three colorimeter measurements around the skin lesion.

Following \cite{kipsColorCorrectionSkin2020}, we use the CIE $\Delta E$ 1976 color difference as the loss function, i.e., the Euclidean distance between predicted and measured colors:

\[
\Delta E =\sqrt{\left(\hat{L}^{*}-L^{*}\right)^{2}+\left(\hat{a}^{*}-a^{*}\right)^{2}+\left(\hat{b}^{*}-b^{*}\right)^{2}},
\]

where \(L^{*},a^{*},b^{*}\) is the ground-truth colorimeter measurement and
\(\hat{L}^{*},\hat{a}^{*},\hat{b}^{*}\) is the predicted skin tone. The estimated ITA can then be calculated using (\ref{eq:ita}).

As in \cite{kipsColorCorrectionSkin2020}, we do not apply white balancing. In our experiments, white balancing reduced skin-tone information (Appendix~\ref{app2}) and is therefore avoided.

The ITA model is trained for up to 50 epochs with early stopping (patience = 5 epochs), using a batch size of 32 and a learning rate of $5 \cdot 10^{-4}$. Because colorimeter measurements are available only for non-lesional sites, fine-tuning is performed exclusively on images of normal skin.

\section{Results}

Models were primarily evaluated on the MSKCC skin tone labeling benchmark using 5-fold cross-validation \cite{weirEvaluatingSkinTone2025}. Splits were performed at the patient level to prevent leakage across folds. We report results for Fitzpatrick type classification and ITA estimation by comparing predictions to reference standards, including agreement with image-based crowdsourced Fitzpatrick labels and with triplicate colorimeter measurements for ITA. Although MSKCC includes both clinical and dermatoscopic images, we report only dermatoscopic results in this section. Finally, we ran inference on commonly used dermatoscopic benchmarks to characterize their skin-tone distributions.

\subsection{Fitzpatrick Type Classification Results}

Across MSKCC, ordinal regression achieved moderate agreement with the in-person expert Fitzpatrick labels, with an overall linear-weighted Cohen’s $\kappa$ of 52.98\% (95\% CI: 51.54, 54.40) (Table~\ref{tab:kappa_by_site_ours_vs_raters}). Agreement varied by anatomical site but remained relatively consistent across the largest strata (anterior torso, posterior torso, upper extremity, lower extremity, head/neck), with $\kappa$ values typically in the low-to-mid 50\% range. Performance decreased for palms/soles, where both the sample size is small, and the site is visually atypical.
Human crowd labels show higher agreement with the expert reference overall (66.08\%; 95\% CI: 64.21, 67.91), which is expected given that crowd labels are produced by multiple raters and can benefit from aggregation.

\begin{table}[htb]
\centering
\small
\setlength{\tabcolsep}{5pt}
\renewcommand{\arraystretch}{1.15}
\caption{Linear-weighted Cohen's $\kappa$ on dermatoscopic images by anatomical site (\% and 95\% CI) compared to human raters.}
\label{tab:kappa_by_site_ours_vs_raters}
\begin{threeparttable}
\begin{tabularx}{\textwidth}{@{}X r >{\raggedright\arraybackslash}X r >{\raggedright\arraybackslash}l@{}}
\toprule
 & \multicolumn{2}{c}{\textbf{Ordinal Regression}} & \multicolumn{2}{c}{\textbf{Crowd\tnote{a}}} \\
\cmidrule(lr){2-3}\cmidrule(lr){4-5}
Site & $n$ & $\kappa$ [\%] (95\% CI) & $n$ & $\kappa$ [\%] (95\% CI) \\
\midrule
anterior torso & 683 & 51.57 (47.84, 55.10) & 430 & 62.24 (57.08, 67.16) \\
head/neck & 301 & 56.04 (51.65, 60.44) & 165 & 63.15 (55.09, 70.66) \\
lateral torso & 30 & 42.15 (21.57, 59.36) & 40 & 71.64 (56.40, 82.68) \\
lower extremity & 778 & 49.18 (45.94, 52.28) & 391 & 68.73 (63.79, 73.11) \\
palms/soles & 26 & 14.57 (5.50, 28.06) & 35 & 54.19 (29.89, 77.46) \\
posterior torso & 928 & 54.96 (52.06, 57.75) & 762 & 66.36 (62.95, 69.61) \\
upper extremity & 932 & 54.30 (51.14, 57.16) & 574 & 60.63 (56.24, 64.79) \\
\midrule
\textbf{Overall} & 3678 & 52.98 (51.54, 54.40) & 2397 & 66.08 (64.21, 67.91) \\
\bottomrule
\end{tabularx}
\begin{tablenotes}[flushleft]
\footnotesize
\item[a] Image-based labels provided by multiple human raters \cite{weirEvaluatingSkinTone2025}.
\end{tablenotes}
\end{threeparttable}
\end{table}

To complement $\kappa$, we report error-based metrics that directly reflect ordinal misclassification severity. Overall, the model achieves a mean average error of 0.84 (95\% CI: 0.72, 0.96) and within-one accuracy of 84.84\% (95\% CI: 79.37, 89.69) (Table~\ref{tab:fp_ordinal_metrics}). The within-one accuracy is higher than crowdsourced annotations (80.77\%; 95\% CI: 77.93, 83.12), indicating that model mistakes are frequently close to the correct Fitzpatrick category even when exact agreement is not achieved. Importantly, the model’s signed error shows low systematic bias overall (Bias = $-0.10$; 95\% CI: $-0.33$, $0.14$), comparable in magnitude to the crowd bias (Bias = $-0.18$; 95\% CI: $-0.26$, $-0.10$). Per-class results suggest that most errors arise from adjacent-category confusions, with the most prominent ambiguity between Fitzpatrick I and II, while the remaining classes show broadly consistent behavior.

\begin{table}[htb]
\centering
\small
\caption{Per-class ordinal regression performance on dermatoscopic images, mean with 95\% CI in parentheses.}
\label{tab:fp_ordinal_metrics}
\begin{threeparttable}
\begin{tabularx}{\textwidth}{@{}X r >{\raggedright\arraybackslash}l >{\raggedright\arraybackslash}l >{\raggedright\arraybackslash}l@{}}
\toprule
Fitzpatrick & $n$ & MAE & Within-one [\%] & Bias \\
\midrule
1 & 598 & 1.19 (1.08, 1.31) & 77.93 (67.95, 85.94) & 1.19 (1.08, 1.31) \\
2 & 713 & 0.51 (0.35, 0.70) & 93.13 (86.54, 98.45) & 0.45 (0.26, 0.66) \\
3 & 595 & 0.72 (0.59, 0.87) & 95.13 (89.00, 98.97) & -0.64 (-0.84, -0.42) \\
4 & 585 & 1.05 (0.82, 1.34) & 75.21 (56.39, 90.26) & -0.40 (-0.99, 0.19) \\
5 & 614 & 0.77 (0.42, 1.22) & 82.57 (64.45, 96.23) & -0.49 (-1.05, -0.02) \\
6 & 573 & 0.85 (0.58, 1.20) & 83.25 (67.09, 94.44) & -0.85 (-1.20, -0.58) \\
\midrule
\textbf{Ours} & 3678 & 0.84 (0.72, 0.96) & 84.83 (79.37, 89.69) & -0.10 (-0.33, 0.14) \\
\textbf{Crowd\tnote{a}} & 2397 & 0.69 (0.63, 0.75) & 80.77 (77.93, 83.12) & -0.18 (-0.26, -0.10) \\
\bottomrule
\end{tabularx}
\begin{tablenotes}[flushleft]
\footnotesize
\item[a] Image-based labels provided by multiple human raters \cite{weirEvaluatingSkinTone2025}.
\end{tablenotes}
\end{threeparttable}
\end{table}

\subsection{ITA Estimation Results}

\begin{figure*}[htb]
\centering
\includegraphics[width=\textwidth]{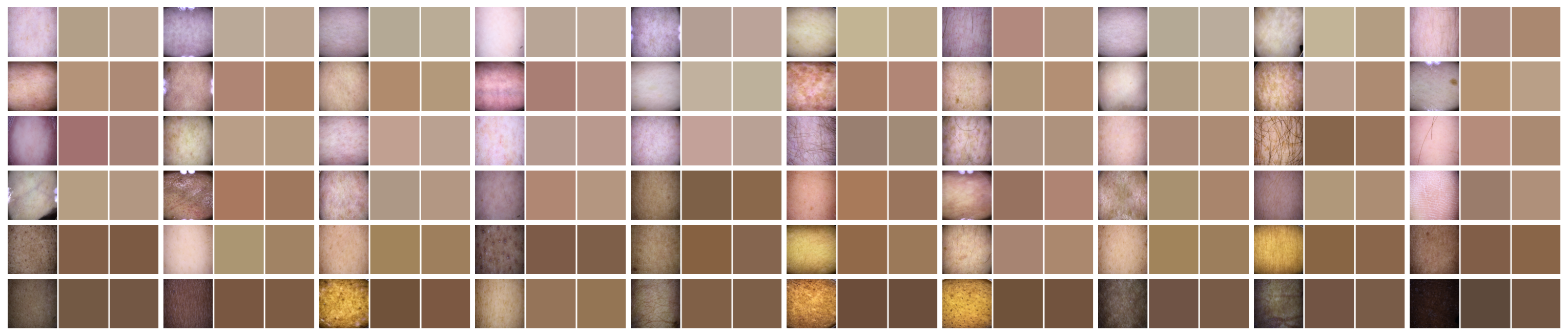}
\caption{Examples of ITA estimation. Each row shows random examples from one Fitzpatrick type from the MSKCC dataset \cite{weirEvaluatingSkinTone2025}. Each entry shows the dermoscopy image (left), the ground-truth skin-color swatch (middle), and the predicted swatch (right), converted from CIELAB to sRGB for visualization.}
\label{fig:ita_triplets}
\end{figure*}

Figure~\ref{fig:ita_triplets} provides qualitative examples of ITA estimation on MSKCC. Across Fitzpatrick strata, predicted skin-color swatches generally match the ground-truth swatches closely, suggesting good visual agreement between predicted and measured skin tone at the lesion-adjacent sampling locations.

Quantitatively, we use ICC3 as the primary agreement metric between the neural network’s predicted ITA and the average triplicate colorimeter ITA (Table~\ref{tab:icc_by_site_sites}). Overall agreement is high, with ICC3 of 93.88\% (95\% CI: 93.31, 94.40) across all sites. Site-wise ICC3 values remain consistently high despite site imbalance in the dataset.

We treat the agreement between the three colorimeter measurements as a theoretical limit of agreement (98.38\%; 95\% CI: 98.24, 98.50). The model's achieved agreement is lower than this theoretical limit (94.12\%, 95\% CI: 93.48, 94.69), but much closer than any baseline method of comparison, including patch-based \cite{bevanDetectingMelanomaFairly2022} and K-means approaches \cite{loaizaSKINTONEPROBLEM2020, bencevicUnderstandingSkinColor2024b} (Table~\ref{tab:icc_overall_methods}).

\begin{table}[htb]
\centering
\small
\caption{ICC3 between neural-network predicted ITA and the average triplicate colorimeter ITA by anatomical site (mean \% and 95\% CI).}
\label{tab:icc_by_site_sites}
\begin{threeparttable}
\begin{tabular}{@{}l r l@{}}
\toprule
Site & $n$ & ICC3 [\%] \\
\midrule
anterior torso & 268 & 94.96 (93.63, 96.01) \\
head/neck & 128 & 95.93 (94.27, 97.11) \\
lower extremity & 373 & 90.17 (88.08, 91.90) \\
posterior torso & 248 & 96.77 (95.87, 97.47) \\
upper extremity & 368 & 95.00 (93.89, 95.91) \\
\bottomrule
\end{tabular}
\end{threeparttable}
\end{table}

\begin{table}[htb]
\centering
\small
\caption{Overall ICC3 for ITA estimation on MSKCC dermatoscopic images ($n = 1385$), comparing our model to computer-vision baselines and the repeatability of triplicate colorimeter measurements (mean \% and 95\% CI).}
\label{tab:icc_overall_methods}
\begin{threeparttable}
\begin{tabular}{@{}l l@{}}
\toprule
Method & ICC3 [\%] \\
\midrule
K-Means \cite{bencevicUnderstandingSkinColor2024b} & 43.64 (39.87, 47.27) \\
Patch-based \cite{bevanDetectingMelanomaFairly2022} & 57.56 (54.42, 60.54) \\
Ours & 94.12 (93.48, 94.69) \\
Colorimeter\tnote{a} & 98.38 (98.24, 98.50) \\
\bottomrule
\end{tabular}
\begin{tablenotes}[flushleft]
\footnotesize
\item[a] ICC3 between three measurements on different locations around the lesion \cite{weirEvaluatingSkinTone2025}.
\end{tablenotes}
\end{threeparttable}
\end{table}

\begin{figure}[htb]
\centering
\includegraphics[width=0.75\textwidth]{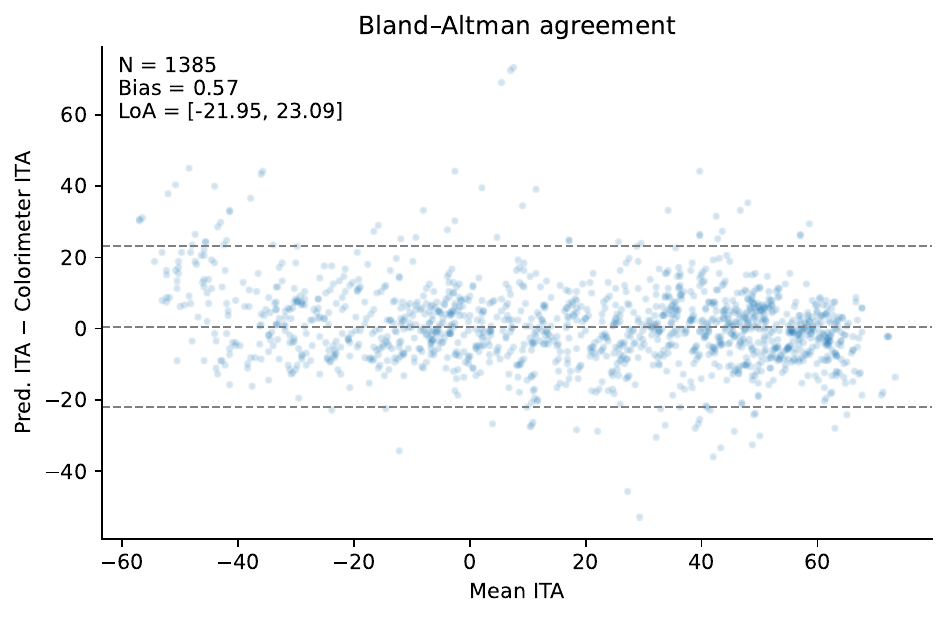}
\caption{Bland--Altman plot for ITA predictions. The solid line shows the mean bias (Predicted -- Colorimeter ITA), and the dashed lines show the 95\% limits of agreement (LoA).}
\label{fig:ita_bland_altman}
\end{figure}

\clearpage

The Bland–Altman analysis in Figure~\ref{fig:ita_bland_altman} shows low systematic bias between predicted and colorimeter ITA and reasonable limits of agreement. The plot indicates that errors are centered near zero with no strong global offset.

\section{Discussion}

In this work, we addressed a key bottleneck in evaluating fairness in dermatoscopy: the lack of reliable skin tone labels in widely used benchmarks. Using the MSKCC skin tone labeling dataset \cite{weirEvaluatingSkinTone2025}, we trained and evaluated an ordinal regression model for Fitzpatrick type estimation and a regression model for ITA estimation under patient-level 5-fold cross-validation. 

Overall, Fitzpatrick type predictions show moderate agreement with the expert in-person reference. The model’s errors are typically small in ordinal distance, and the overall bias is close to zero. While the agreement of our model is lower than the best-case agreement achievable by aggregating multiple human raters, it still supports the model’s practical use as a fast, scalable tool for annotating large dermatoscopic datasets where manual labeling would be slow and costly.

Perhaps more importantly, our ITA estimation achieves high agreement with triplicate colorimeter measurements (Table~\ref{tab:icc_by_site_sites}). This result is notable in light of prior evidence that pixel-based skin-tone estimates are not feasible due to illumination variation, camera pipelines, and local tissue effects \cite{weirEvaluatingSkinTone2025}. In the overall comparison, our model substantially outperforms previously used computer-vision baselines for ITA estimation (Table~\ref{tab:icc_overall_methods}), being much closer to the theoretical limit of agreement, i.e., between multiple colorimeter measurements. Taken together, these results suggest that learning-based approaches can recover clinically relevant skin tone signals from dermatoscopic images far more effectively than earlier strategies based on averaging pixel values.

\subsection{Skin Tone Distribution of Commonly Used Dermatoscopic Benchmarks}

To contextualize model performance and characterize common dermatoscopic benchmarks, we ran inference on ISIC 2020 \cite{rotembergPatientcentricDatasetImages2021a} and MILK10k \cite{tschandlMILK10kHierarchicalMultimodal2026}. For the Fitzpatrick type, we aggregated predictions via majority vote across the five folds; for ITA, we averaged the fold-wise CIELAB predictions and then computed ITA. We compare these inferred distributions to MSKCC \cite{weirEvaluatingSkinTone2025}, where both Fitzpatrick type and ITA are available as ground truth, and the dataset is comparatively balanced across skin tones.

\begin{figure}[htb]
\centering
\includegraphics[width=0.75\textwidth]{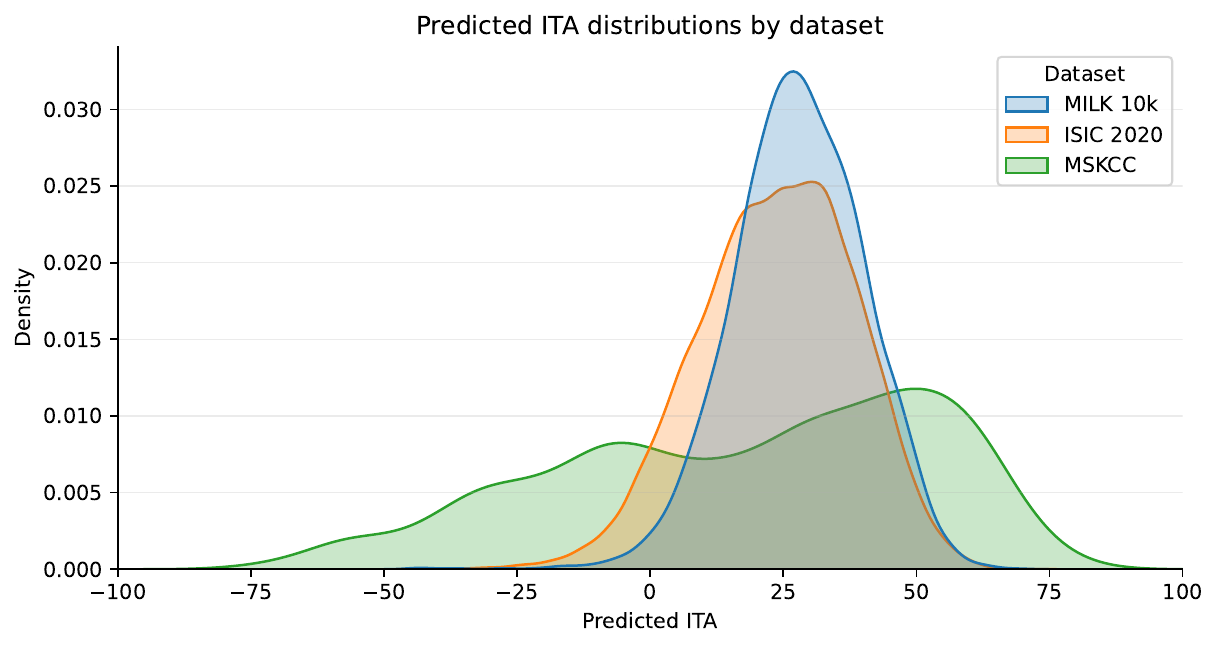}
\caption{Distribution of ITA values across datasets. For ISIC 2020 \cite{rotembergPatientcentricDatasetImages2021a} and MILK10k \cite{tschandlMILK10kHierarchicalMultimodal2026}, ITA values are model predictions; for MSKCC \cite{weirEvaluatingSkinTone2025}, ITA values correspond to ground-truth measurements.}
\label{fig:lab_inference}
\end{figure}

Figure~\ref{fig:lab_inference} shows the resulting ITA distributions. ISIC 2020 and MILK10k concentrate strongly in the mid-range (predominantly between 0 and 50 ITA). While MILK10k appears slightly more diverse, there are still virtually no samples below approximately $-25$ ITA. In contrast, MSKCC spans a broader range of measured ITA values (approximately $-50$ to $75$), providing substantially wider skin-tone coverage.

\begin{figure}[htb]
\centering
\includegraphics[width=\textwidth]{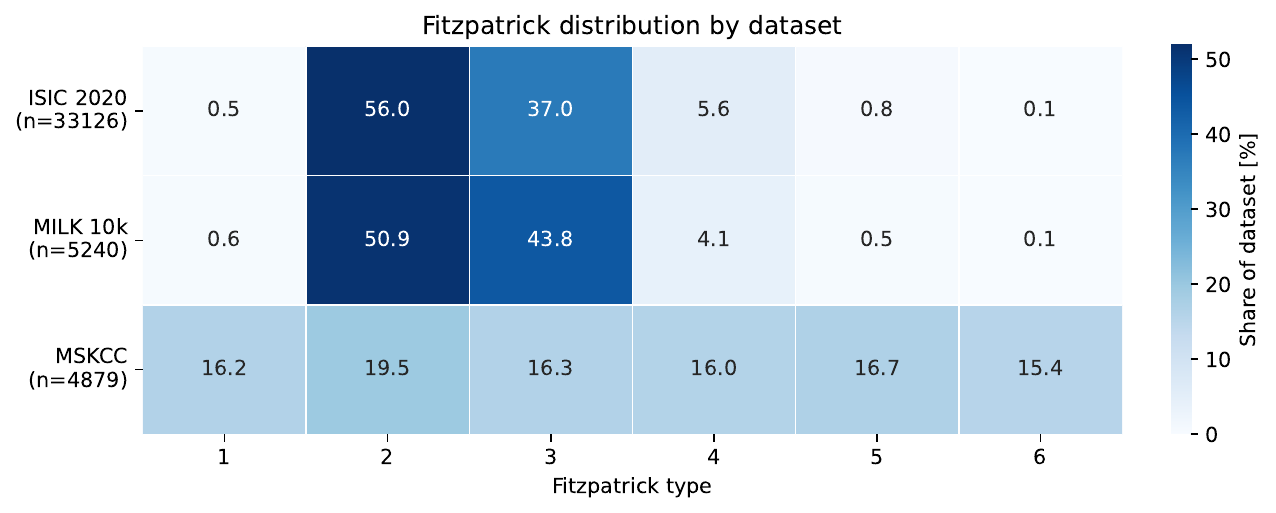}
\caption{Heatmap of Fitzpatrick skin type distributions across datasets (within-dataset percentages). For ISIC 2020 \cite{rotembergPatientcentricDatasetImages2021a} and MILK10k \cite{tschandlMILK10kHierarchicalMultimodal2026}, Fitzpatrick types are model predictions; for MSKCC \cite{weirEvaluatingSkinTone2025}, ground-truth Fitzpatrick labels are shown.}
\label{fig:fp_inference}
\end{figure}

The same lack of diversity is apparent when examining Fitzpatrick type composition (Figure~\ref{fig:fp_inference}). ISIC 2020 (n=33{,}126) and MILK10k (n=5{,}240) are dominated by Fitzpatrick II and III, with only 0.5--0.6\% predicted as Fitzpatrick I and 0.9\% (ISIC 2020) / 0.6\% (MILK10k) predicted as Fitzpatrick V--VI combined. Even given that our classifier has a known tendency to confuse Fitzpatrick I with II, the near-absence of Fitzpatrick I and the extremely small share of Fitzpatrick VI and V indicate that these benchmarks are inadequate to determine model performance of the full range of skin tones.

\subsection{Limitations}

This study has two key limitations. Firstly, although we use patient-level 5-fold cross-validation, it is difficult to quantify how well the approach generalizes to unseen acquisition settings because no comparable dermatoscopy dataset with in-person Fitzpatrick labels and triplicate colorimeter measurements is currently available. The MSKCC data are benign-only and originate from a single center \cite{weirEvaluatingSkinTone2025}, which may not capture the full variability of devices, imaging protocols, geographic populations, and clinical workflows encountered in practice.

Secondly, the MSKCC dataset contains both lesional and non-lesional (normal skin) images. The ITA estimation, in particular, was only evaluated on non-lesional sites since only those have available colorimeter measurements. Nevertheless, we believe the results are still a good proxy for how well the method would generalize to lesional sites, especially if combined with lesion segmentation as a preprocessing step to remove the lesion region from the image.

Despite these limitations, the method has immediate practical value as a scalable labeling tool. By enabling consistent estimation of Fitzpatrick type and ITA on large dermatoscopic datasets where skin tone labels do not exist, it can support more rigorous audits of subgroup performance, dataset composition, and fairness-related failure modes. We release the approach as an open-source tool to facilitate reproducible skin tone characterization in dermatoscopy and to lower the barrier for fairness evaluation in future work.

\section*{Acknowledgements}

\subsection*{Declaration of Funding}

This research was funded by the European Union-NextGenerationEU.

\subsection*{Declaration of generative AI and AI-assisted technologies in the manuscript preparation process}

During the preparation of this work the authors used OpenAI ChatGPT in order to prepare and format tables as well as perform proofreading. After using this service, the authors reviewed and edited the content as needed and take full responsibility for the content of the published article.

\subsection*{Declaration of Competing Interest}

The authors declare no conflict of interest.

\subsection*{Declaration of Ethical Approval}

No human or animal subjects have been part of this study.

\clearpage

\appendix
\section{Comparison of Ordinal Regression to Classification}
\label{app1}

To estimate Fitzpatrick types, we have also trained a standard classification model to use as a baseline for our ordinal regression approach. The model is trained in the same way as the ordinal regression model, but instead of using CORAL, it uses a regular classification head. It is trained using cross-entropy loss using the same hyperparameters as the ordinal regression network. The results of testing the model using 5-fold cross-validation are shown in Table~\ref{tab:kappa_by_site_ours_vs_class}.

\begin{table}[h]
\centering
\small
\setlength{\tabcolsep}{5pt}
\renewcommand{\arraystretch}{1.15}
\caption{Linear-weighted Cohen's $\kappa$ by anatomical site (\% and 95\% CI): ordinal regression vs a naive classification model. Results are shown for both clinical close-up and dermatoscopic images.}
\label{tab:kappa_by_site_ours_vs_class}
\begin{tabular}{@{}l r l l@{}}
\toprule
 &  & \textbf{Ordinal Regression} & \textbf{Classification} \\
\cmidrule(lr){3-3}\cmidrule(lr){4-4}
Site & Images ($n$) & $\kappa$ [\%] (95\% CI) & $\kappa$ [\%] (95\% CI) \\
\midrule
anterior torso & 907 & 52.11 (49.01, 55.09) & 48.10 (43.57, 52.35) \\
head/neck & 398 & 55.16 (51.14, 59.07) & 39.12 (32.48, 45.58) \\
lateral torso & 40 & 42.90 (26.90, 58.07) & 38.24 (15.29, 64.49) \\
lower extremity & 1033 & 49.52 (46.75, 52.32) & 40.71 (36.40, 45.13) \\
palms/soles & 35 & 15.19 (7.26, 27.14) & 14.48 (6.53, 28.00) \\
posterior torso & 1231 & 54.45 (51.87, 57.10) & 46.69 (42.86, 50.31) \\
upper extremity & 1235 & 54.29 (51.64, 56.71) & 48.76 (44.82, 52.17) \\
\midrule
\textbf{Overall} & 4879 & 53.00 (51.70, 54.22) & 45.39 (43.42, 47.32) \\
\bottomrule
\end{tabular}
\end{table}

\clearpage

\section{Impact of White Balancing on Skin Tone Estimation}
\label{app2}

In our experiments, applying Shades-of-Gray white balancing disproportionately affects Fitzpatrick types I, II, V, and VI (Table~\ref{tab:bias_wb}). A plausible explanation is that Shades-of-Gray estimates the scene illuminant from global image statistics; in dermatoscopic images, these statistics are strongly influenced by the fraction of visible skin, the lesion itself, and acquisition artifacts. Thus, the resulting color correction can shift skin pixels in a skin-tone–dependent manner, altering downstream skin-tone estimates. For this reason, we do not recommend white balancing as a preprocessing step for dermatoscopic skin tone estimation.

\begin{table}[h]
\centering
\small
\setlength{\tabcolsep}{5pt}
\renewcommand{\arraystretch}{1.15}
\caption{Per-class bias, i.e., average signed error (mean and 95\% CI): Ordinal regression without white-balancing vs. with Shades of Gray white balancing. Linear-weighted Cohen's $\kappa$ [\%] (mean and 95\% CI) is shown on the last row. Results are shown for both clinical close-up and dermatoscopic images.}
\label{tab:bias_wb}
\begin{threeparttable}
\begin{tabular}{@{}l r l l@{}}
\toprule
FP & $n$ & No White Balance & Shades of Gray \\
\midrule
1 & 789 & 1.21 (1.17, 1.25) & 1.28 (1.23, 1.32) \\
2 & 950 & 0.47 (0.43, 0.52) & 0.56 (0.52, 0.61) \\
3 & 794 & -0.57 (-0.62, -0.52) & -0.40 (-0.45, -0.34) \\
4 & 783 & -0.33 (-0.42, -0.25) & -0.29 (-0.37, -0.21) \\
5 & 813 & -0.43 (-0.51, -0.35) & -0.52 (-0.59, -0.45) \\
6 & 750 & -0.83 (-0.89, -0.77) & -0.99 (-1.04, -0.93) \\
\midrule
\textbf{All} & 4879 & -0.06 (-0.09, -0.03) & -0.04 (-0.07, -0.00) \\
\midrule
\textbf{$\kappa$} & & 53.00 (51.70, 54.22) & 50.72 (49.11, 52.24) \\
\bottomrule
\end{tabular}
\end{threeparttable}
\end{table}

\clearpage

\bibliographystyle{elsarticle-num-names}
\bibliography{bibliography.bib}



\end{document}